\documentclass[10pt,twocolumn,letterpaper,fleqn]{article}
\usepackage[title]{appendix}

\usepackage{cvpr}
\usepackage{float}
\usepackage{times}
\usepackage{epsfig}
\usepackage{graphicx}
\usepackage{amsmath}
\usepackage{amssymb}
\usepackage{multirow}

\makeatletter
\def\and{%                  % \begin{tabular}[t]{c}
  \end{tabular}%
  \hskip 1em \@plus.17fil%
  \begin{tabular}[t]{c}}%   % \end{tabular}
\makeatother 

\usepackage[pagebackref=true,breaklinks=true,letterpaper=true,colorlinks,bookmarks=false]{hyperref}

\cvprfinalcopy % *** Uncomment this line for the final submission

\def\cvprPaperID{0000} % *** Enter the CVPR Paper ID here

\ifcvprfinal\fi
\begin{document}

%%%%%%%%% TITLE
\title{f-BRS: Rethinking Backpropagating Refinement for Interactive Segmentation}

\author{Konstantin Sofiiuk\\
\and
Ilia Petrov\\
\and
Olga Barinova\\
\and
Anton Konushin\\
\and
{\tt\small {\{\href{mailto:k.sofiiuk@samsung.com}{k.sofiiuk}, \href{mailto:ilia.petrov@samsung.com}{ilia.petrov}, \href{mailto:o.barinova@samsung.com}{o.barinova}, \href{mailto:a.konushin@samsung.com}{a.konushin}\}@samsung.com}}\\
Samsung AI Center -- Moscow\\
}

\maketitle
\ifcvprfinal\thispagestyle{empty}\fi

%%%%%%%%% ABSTRACT
\begin{abstract}
Deep neural networks have become a mainstream approach to interactive segmentation. As we show in our experiments, while for some images a trained network provides accurate segmentation result with just a few clicks, for some unknown objects it cannot achieve satisfactory result even with a large amount of user input. Recently proposed backpropagating refinement scheme (BRS) \cite{jang2019interactive} introduces an optimization problem for interactive segmentation that results in significantly better performance for the hard cases. At the same time, BRS requires running forward and backward pass through a deep network several times that leads to significantly increased computational budget per click compared to other methods. We propose f-BRS (feature backpropagating refinement scheme) that solves an optimization problem with respect to auxiliary variables instead of the network inputs, and requires running forward and backward passes just for a small part of a network. Experiments on GrabCut, Berkeley, DAVIS and SBD datasets set new state-of-the-art at an order of magnitude lower time per click compared to original BRS \cite{jang2019interactive}. The code and trained models are available at \url{https://github.com/saic-vul/fbrs_interactive_segmentation}.
\end{abstract}

%%%%%%%%% BODY TEXT
\section{Introduction}

The development of robust models for visual understanding is tightly coupled with data annotation. For instance, one self-driving car can produce about 1Tb of data every day. Due to constant changes in environment new data should be annotated regularly. 

Object segmentation provides fine-grained scene representation and can be useful in many applications, \eg autonomous driving, robotics, medical image analysis, etc. However, practical use of object segmentation is now limited due to extremely high annotation costs. Several large segmentation benchmarks \cite{benenson2019large, gupta2019lvis} with millions of annotated object instances came out recently. Annotation of these datasets became feasible with the use of automated interactive segmentation methods \cite{agustsson2019interactive, benenson2019large}.

Interactive segmentation has been a topic of research for a long time \cite{rother2004grabcut, grady2006random, gulshan2010geodesic, hariharan2011semantic, bai2014error, xu2016deep, li2018interactive, maninis2018deep, jang2019interactive}. The main scenario considered in the papers is click-based segmentation when the user provides input in a form of positive and negative clicks. Classical approaches formulate this task as an optimization problem \cite{boykov2001interactive, grady2006random, gulshan2010geodesic, hariharan2011semantic, bai2014error}. These methods have many built-in heuristics and do not use semantic priors to full extent, thus requiring a large amount of input from the user. On the other hand, deep learning-based methods \cite{xu2016deep, li2018interactive, maninis2018deep} tend to overuse image semantics. While showing great results on the objects that were present in the training set, they tend to perform poorly on unseen object classes. Recent works propose different solutions to these problems \cite{liew2017regional, li2018interactive, majumder2019content}. Still, state-of-the-art networks for interactive segmentation are either able to accurately segment the object of interest after a few clicks or do not provide satisfactory result after any reasonable number of clicks (see Section \ref{sec:convergence} for experiments).

The recently proposed backpropagating refinement scheme (BRS) \cite{jang2019interactive} brings together optimization-based and deep learning-based approaches to interactive segmentation. BRS enforces the consistency of the resulting object mask with user-provided clicks. The effect of BRS is based on the fact that small perturbations of the inputs for a deep network can cause massive changes in the network output \cite{szegedy2013intriguing}. Thus, BRS requires running forward and backward pass multiple times through the whole model, which substantially increases computational budget per click compared to other methods and is not practical for many end-user scenarios.

\begin{figure*}[!ht]
 \centering
 \includegraphics[width=0.95\linewidth]{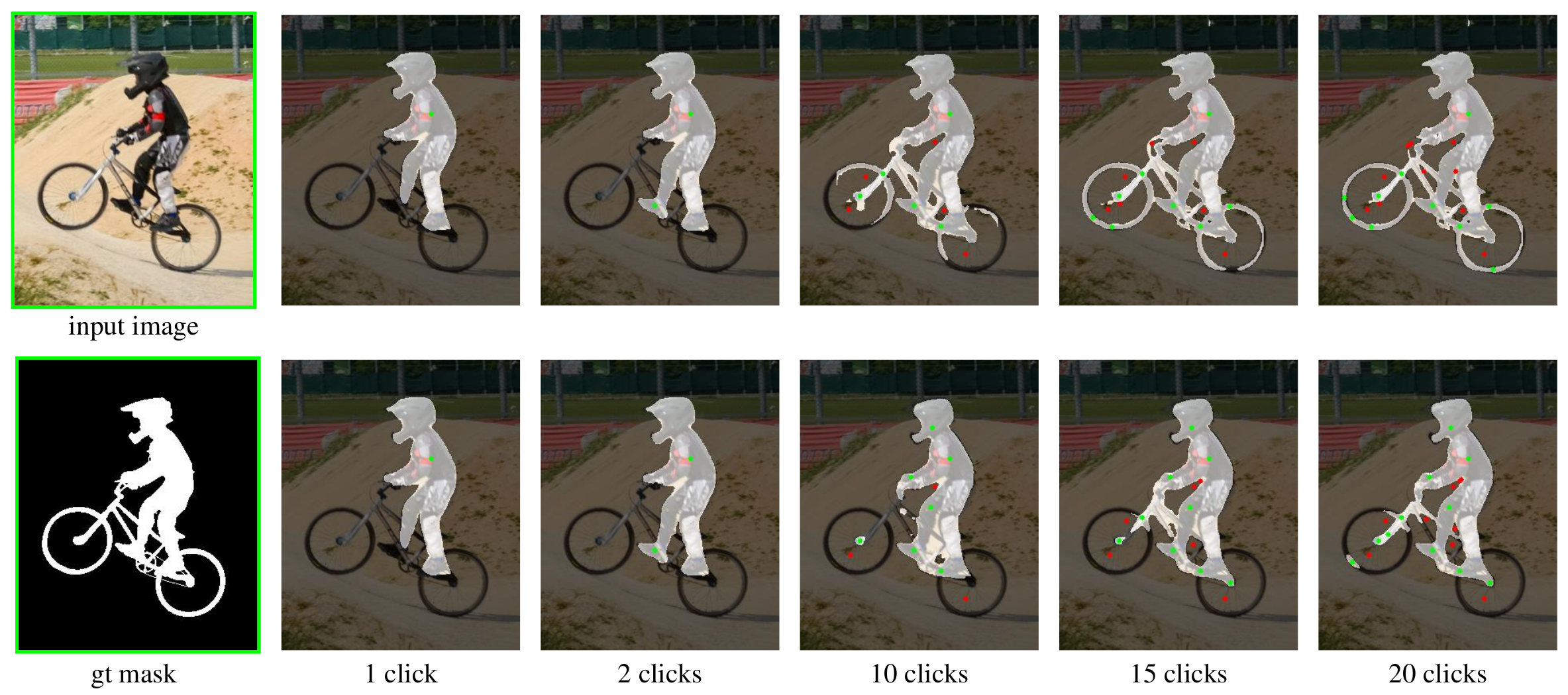}
 \caption{Results of interactive segmentation on an image from DAVIS dataset. First row: using proposed f-BRS-B (Section \ref{sec:proposed}), second row: without BRS. Green dots denote positive clicks, red dots denote negative clicks.}   
 \label{fig:nice_brs}
\end{figure*}

In this work we propose \emph{f-BRS (feature backpropagating refinement scheme)} that reparameterizes the optimization problem and thus requires running forward and backward passes only through a small part of the network (\ie last several layers). Straightforward optimization for activations in a small sub-network would not lead to the desired effect because the receptive field of the convolutions in the last layers relative to the output is too small. Thus we introduce a set of auxiliary parameters for optimization that are invariant to the position in the image. We show that optimization with respect to these parameters leads to a similar effect as the original BRS, without the need to compute backward pass  through the whole network. 

We perform experiments on standard datasets: GrabCut \cite{rother2004grabcut}, Berkeley \cite{martin2001database}, DAVIS \cite{perazzi2016benchmark} and SBD \cite{hariharan2011semantic}, and show state-of-the-art results, improving over existing approaches in terms of speed and accuracy.

\section{Related work}

%\subsubsection{Interactive segmentation}

The goal of interactive image segmentation is to obtain an accurate mask of an object using minimal user input. Most of the methods assume such interface where a user can provide positive and negative clicks (seeds) several times until the desired object mask is obtained. 

\textbf{Optimization-based methods.} Before deep learning, interactive segmentation was usually posed as an optimization problem. Li \etal \cite{li2004lazy} used graph-cut algorithm to separate foreground pixels from the background using distances from each pixel to foreground and background seeds in color space. Grady \etal \cite{grady2006random} proposed a method based on random walks, where each pixel is marked according to the label of the first seed that the walker reaches. Later, \cite{gulshan2010geodesic} compute geodesic distances from the clicked points to every image pixel and use them in energy minimisation. In \cite{kim2010nonparametric}, several segmentation maps are first generated for an image. Then optimization algorithm is applied to the cost function that enforces pixels of the same segment to have the same label in the resulting segmentation mask. 

Optimization-based methods usually demonstrate predictable behaviour and allow obtaining detailed segmentation masks with enough user input. Since no learning is involved, the amount of input required from a user does not depend on the type of an object of interest. The main drawback of this approach is insufficient use of semantic priors. This requires additional user effort to obtain accurate object masks for known objects compared to recently proposed learning-based methods.

\textbf{Learning-based methods.} The first deep learning-based interactive segmentation method was proposed in \cite{xu2016deep}. They calculate distance maps from positive and negative clicks, stack them together with an input image and pass into a network that predicts an object mask. This approach was later used in most of the following works. Liew \etal \cite{liew2017regional} propose to combine local predictions on patches containing user clicks and thus refine network output. Li \etal \cite{li2018interactive} notice that learnt models tend to be overconfident in their predictions. In order to improve diversity of the outputs, they generate multiple masks and then select one among them. In \cite{song2018seednet}, user annotations are multiplied automatically by locating foreground and background clicks. 

The common problem of all deep-learning-based methods for interactive segmentation is overweighting semantics and making little use of user-provided clicks. This happens because during training user clicks are in perfect correspondence with the semantics of the image and add little information, therefore can be easily downweighted during the training process.

\begin{figure*}[t]
    \centering
    \includegraphics[width=0.9\linewidth]{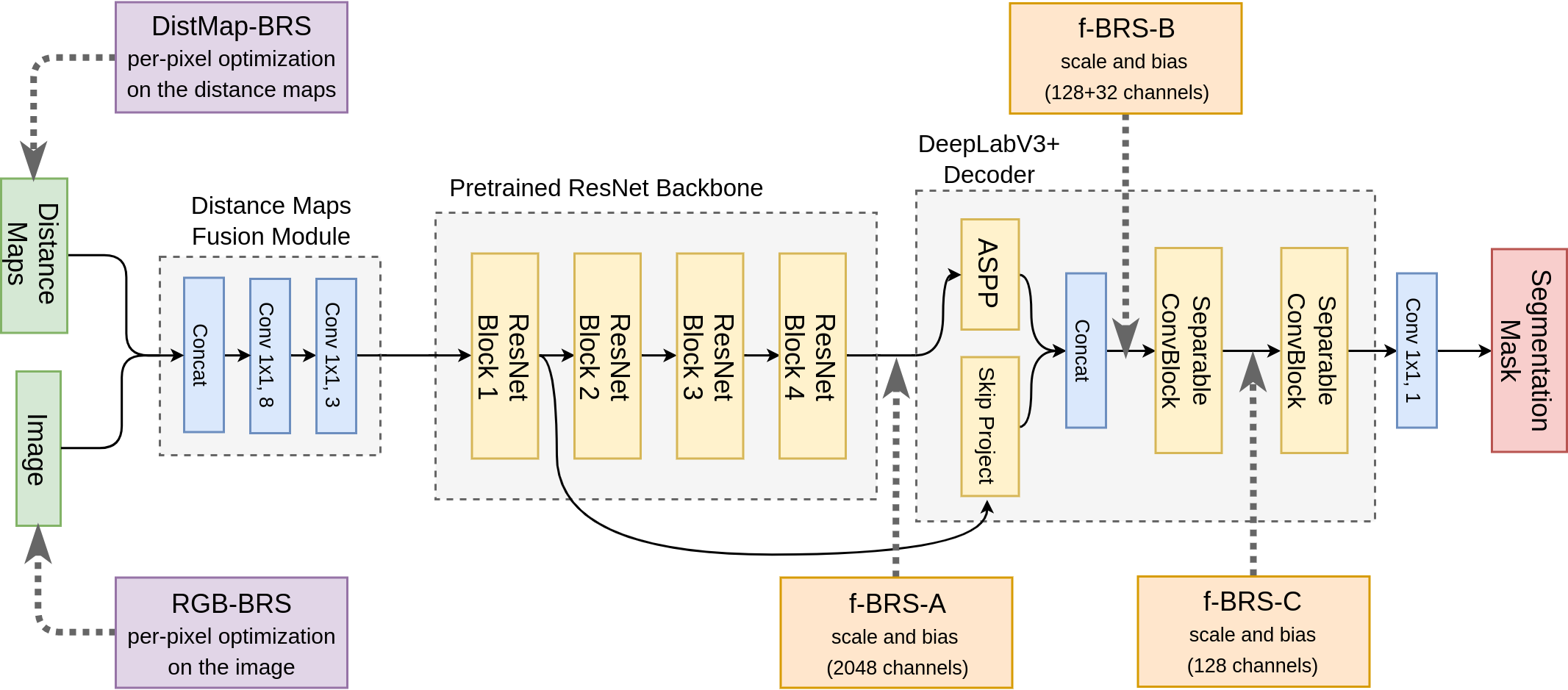}
    \caption{Illustration of the proposed method described in Section \ref{sec:proposed}. f-BRS-A optimizes scale and bias for the features after pre-trained backbone, f-BRS-B optimizes scale and bias for the features after ASPP, f-BRS-C optimizes scale and bias for the features after the first separable convblock. The number of channels is provided for ResNet-50 backbone.}
    \label{fig:f_brs}
\end{figure*}

\textbf{Optimization for activations.} Optimization schemes that update activation responses while keeping weights of a neural network fixed have been used for different problems \cite{simonyan2013deep, yan2013hierarchical, zhang2018top, gatys2015texture, gatys2016image}. Szegedy \etal \cite{szegedy2013intriguing} formulated an optimization problem for generating adversarial examples, \ie images that are visually indistinguishable from the natural ones, though are incorrectly classified by the network with high confidence. They demonstrated that in deep networks small perturbation of an input signal can cause large changes in activations of the last layers. In \cite{jang2019interactive}, the authors apply this idea to the problem of interactive segmentation. They find minimal edits to the input distance maps that result in an object mask consistent with user-provided annotation. 

In this work, we also formulate an optimization problem for interactive segmentation. In contrast to \cite{jang2019interactive}, here we do not perform optimization over the network inputs but introduce an auxiliary set of parameters for optimization. After such reparameterization, we do not need to run forward and backward passes through the whole network. We evaluate different reparameterizations and the speed and accuracy of the resulting methods. The derived optimization algorithm f-BRS is an order of magnitude faster than BRS from \cite{jang2019interactive}.

%-------------------------------------------------------------------------

\section{Proposed method}
\label{sec:proposed}

First, let us recall optimization problems from the literature, where the fixed network weights were used. Below we use a unified notation. We denote the space of input images by $\mathbb{R}^m$ and a function that a deep neural network implements by $f$. 

\subsection{Background}

\textbf{Adversarial examples generation.}
Szegedy \etal \cite{szegedy2013intriguing} formulate an optimization problem for generating adversarial examples for an image classification task. They find images that are visually indistinguishable from the natural ones, which are incorrectly classified by the network. Let $\mathcal{L}$ denote a continuous loss function that penalizes for incorrect classification of an image. For a given image $x \in R^m$ and target label $l \in \{1 \dots k\}$, they aim to find $x + \Delta x$, which is the closest image to $x$ classified as $l$ by $f$. For that they solve the following optimization problem:
\begingroup
\setlength\abovedisplayskip{6pt}
\setlength\belowdisplayskip{6pt}
\begin{flalign}
\begin{split}
||\Delta x&||_2 \rightarrow \min \quad \text{subject to} \\
\text{1.}\,  &f(x + \Delta x) = l \\
\text{2.}\,  &x + \Delta x \in [0, 1]^m 
\label{eq:adversarial}
\end{split}
\end{flalign}
\endgroup

This problem in (\ref{eq:adversarial}) is reduced to minimisation of the following energy function:
\begingroup
\setlength\abovedisplayskip{6pt}
\setlength\belowdisplayskip{6pt}
\begin{flalign}
\label{eq:adversarial_reduced}
\lambda||\Delta x||_2 + \mathcal{L}(f(x + \Delta x), l) \rightarrow \min_{\Delta x}
\end{flalign}
\endgroup

The variable $\lambda$ in later works is usually assumed a constant and serves as a trade-off between the two energy terms.  \\[3pt]
 
\textbf{Backpropagating refinement scheme for interactive segmentation.}
Jang \etal \cite{jang2019interactive} propose a backpropagating refinement scheme that applies a similar optimization technique to the problem of interactive image segmentation. In their work, a network takes as input an image stacked together with distance maps for user-provided clicks. They find minimal edits to the distance maps that result in an object mask consistent with user-provided annotation. For that, they minimise a sum of two energy functions, \ie corrective energy and inertial energy. Corrective energy function enforces consistency of the resulting mask with user-provided annotation, and inertial energy prevents excessive perturbations in the network inputs.

Let us denote the coordinates of user-provided click by $(u, v)$ and its label (positive or negative) as $l \in \{0, 1\}$. Let us denote the output of a network $f$ for an image $x$ in position $(u, v)$ as $f(x)_{u, v}$ and the set of all user-provided clicks as $\{(u_i, v_i, l_i)\}_{i = 1}^n$. The optimization problem in \cite{jang2019interactive} is formulated as follows:
\begingroup
\setlength\abovedisplayskip{6pt}
\setlength\belowdisplayskip{6pt}
\setlength{\mathindent}{0cm}
\begin{equation}
\label{eq:brs_optimization}
\quad \lambda||\Delta x||_2 + \sum_{i=1}^n\big(f(x + \Delta x)_{u_i, v_i} - l_i\big)^2 \rightarrow \min_{\Delta x},
\end{equation}
\endgroup
where the first term represents inertial energy, the second term represents corrective energy and $\lambda$ is a constant that regulates trade-off between the two energy terms. This optimization problem resembles the one from (\ref{eq:adversarial_reduced}) with classification loss for one particular label replaced by a sum of losses for the labels of all user-provided clicks. Here we do not need to ensure that the result of optimization is a valid image, so the energy (\ref{eq:brs_optimization}) can be minimised by unconstrained L-BFGS.

The main drawback of this approach is that L-BFGS requires computation of gradients with respect to network inputs, \ie backpropagating through the whole network.  It is computationally expensive and results in significant computational overhead.

We also notice that since the first layer of a network $f$ is a convolution, \ie a linear combination of the inputs, one can minimise the energy (\ref{eq:brs_optimization}) with respect to input image instead of distance maps and obtain equivalent solution. Moreover, if we minimise it with respect to RGB image which is invariant to an interactive input, we can use the result as an initialisation for optimization of (\ref{eq:brs_optimization}) with new clicks. Thus, we set the BRS with respect to an input image as a baseline in our experiments and denote it as RGB-BRS. For a fair comparison, we also implement the optimization with respect to the input distance maps (DistMap-BRS), that was originally introduced in \cite{jang2019interactive}.

\subsection{Feature backpropagating refinement}

In order to speed-up the optimization process, we want to compute backpropagation not for the whole network, but for some part of it. This can be achieved by optimizing some intermediate parameters in the network instead of the input. A naive approach would be to simply optimize the outputs of some of the last layers and thus compute backpropagation only through the head of a network. However, such a naive approach would not lead to the desired result. The convolutions in the last layers have a very small receptive field with respect to the network outputs. Therefore, an optimization target can be easily achieved by changing just a few components of a feature tensor which would cause only minor localized changes around the clicked points in the resulting object mask.

Let us reparameterize the function $f$ and introduce auxiliary variables for optimization. Let  $\hat{f}(x, z)$ denote the function that depends both on the input $x$ and on the introduced variables $z$. With auxiliary parameters fixed $z=p$ the reparameterized function is equivalent to the original one $\hat{f}(x, p) \equiv f(x)$. Thus, we aim to find a small value of $\Delta p$, which would bring the values of $\hat{f}(x, p + \Delta p)$ in the clicked points close to the user-provided labels. We formulate the optimization problem as follows:
\begingroup
\setlength\abovedisplayskip{6pt}
\setlength\belowdisplayskip{6pt}
\setlength{\mathindent}{0cm}
\begin{flalign}
\label{eq:fbrs_optimization}
\quad \lambda||\Delta p||_2 + \sum_{i=1}^n\big(\hat{f}(x, p + \Delta p)_{u_i, v_i} - l_i\big)^2 \rightarrow \min_{\Delta p}.
\end{flalign}
\endgroup

We call this optimization task \emph{f-BRS (feature backpropagating refinement)} and use unconstrained L-BFGS optimizer for minimization. For f-BRS to be efficient, we need to choose the reparameterization that a) does not have a localized effect on the outputs, b) does not require a backward pass through the whole network for optimization. 

One of the options for such reparameterization may be channel-wise scaling and bias for the activations of the last layers in the network. Scale and bias are invariant to the position in the image, thus changes in this parameters would affect the results globally. Compared to optimization with respect to activations, optimization with respect to scale and bias cannot result in degenerate solutions (\ie minor localized changes around the clicked points). 

Let us denote the output of some intermediate layer of the network for an image $x$ by $F(x)$, the number of its channels by $h$, a function that the network head implements by $g$. Thus, $f$ can be represented by $f(x) \equiv g(F(x))$. Then the reparameterized function $\hat{f}$ looks as follows:
\begingroup
\setlength\abovedisplayskip{6pt}
\setlength\belowdisplayskip{6pt}
\begin{flalign}
\label{eq:scale_bias}
\hat{f}(x, s, b) = g\big(s \cdot F(x) + b\big), 
\end{flalign}
\endgroup
where $b \in \mathbb{R}^h$ is a vector of biases, $s \in \mathbb{R}^h$ is a vector of scaling coefficients and $\cdot$ denotes a channel-wise multiplication. For $s=\mathbf{1}$ and $b=\mathbf{0}$ we have $\hat{f}(x) \equiv f$, thus we take these values as initial values for optimization.

By varying the part of the network to which auxiliary scale and bias are applied, we achieve a natural trade-off between accuracy and speed. Figure \ref{fig:f_brs} shows the architecture of the network that we used in this work and illustrates different options for optimization. Surprisingly, we found that applying f-BRS to the last several layers causes just a small drop of accuracy compared to full-network BRS, and leads to significant speed-up.

\section{Zoom-In for interactive segmentation}
\label{sec:zoomin}

Previous works on interactive segmentation often used inference on image crops to achieve speed-up and preserve fine details in the segmentation mask. Cropping helps to infer the masks of small objects, but it also may degrade results in cases when an object of interest is too large to fit into one crop. 

In this work, we use an alternative technique (we call it \emph{Zoom-In}), which is quite simple but improves both quality and speed of the interactive segmentation. It is based on the ideas from object detection \cite{lu2016adaptive, gao2018dynamic}. We have not found any mentions of this exact technique in the literature in the context of interactive segmentation, so we describe it below.

We noticed that the first 1-3 clicks are enough for the network to achieve around 80\% IoU with ground truth mask in most cases. It allows us to obtain a rough crop around the region of interest. Therefore, starting from the third click we crop an image according to the bounding box of the inferred object mask and apply the interactive segmentation only to this Zoom-In area. We extend the bounding box by 40\% along sides in order to preserve the context and not miss fine details on the boundary. If a user provides a click outside the bounding box, we expand or narrow down the zoom-in area. Then we resize the bounding box so that its longest side matches 400 pixels. Figure \ref{fig:zoomin} shows an example of Zoom-In.

This technique helps the network to predict more accurate masks for small objects. In our experiments, Zoom-In consistently improved the results, therefore we used it by default in all experiments in this work. Table \ref{tab:comparison_zoomin} shows a quantitative comparison of the results with and without Zoom-In on GrabCut and Berkeley datasets.

\begingroup
\begin{figure}
    \centering
    \includegraphics[width=\linewidth]{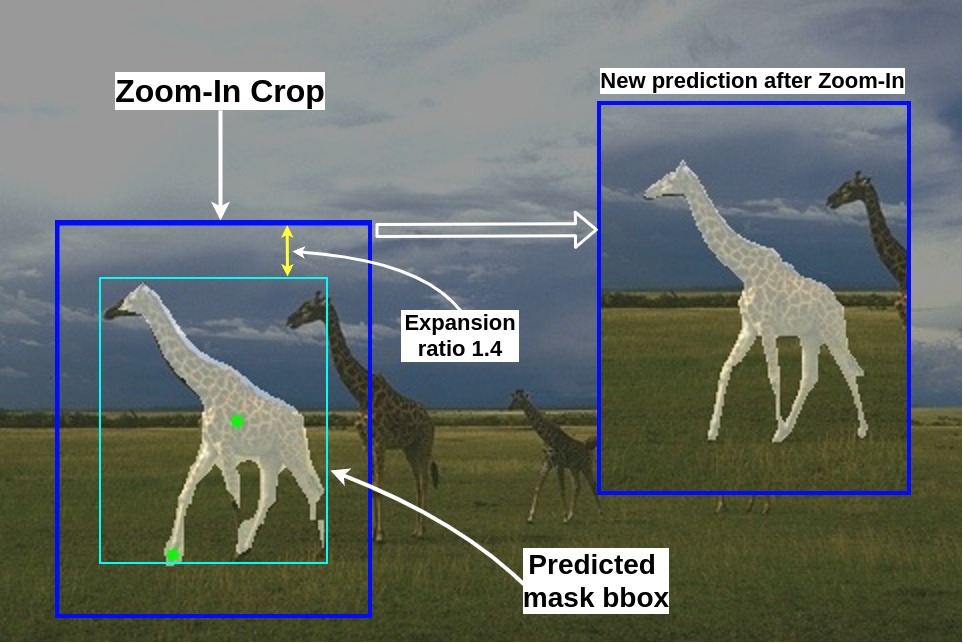}
    \caption{Example of applying zoom-in technique described in Section \ref{sec:zoomin}. See how cropping an image allows recovering fine details in the segmentation mask.}
    \label{fig:zoomin}
\end{figure}
\endgroup

\section{Experiments}
\label{sec:experiments}

Following the standard experimental protocol, we evaluate proposed method on the following datasets: SBD \cite{hariharan2011semantic}, GrabCut \cite{rother2004grabcut}, Berkeley \cite{martin2001database} and DAVIS \cite{perazzi2016benchmark}. 

\textbf{GrabCut.} The GrabCut dataset contains 50 images with a single object mask for each image.

\textbf{Berkeley.} For the Berkeley dataset, we use the same test set as in \cite{mcguinness2010comparative}, which includes 96 images with 100 object masks for testing. 

\textbf{DAVIS.} The DAVIS dataset is used for evaluating video segmentation algorithms. To evaluate interactive segmentation algorithms one can sample random frames from the videos. We use the same 345 individual frames from video sequences as \cite{jang2019interactive} for evaluation. To follow the evaluation protocol we combine instance-level object masks into one semantic segmentation mask for each image.

\textbf{SBD.} The SBD dataset was first used for evaluating object segmentation techniques in \cite{xu2016deep}. The dataset contains 8,498 training images and 2,820 test images. As in previous works, we train the models on the training part and use the validation set, which includes 6,671 instance-level object masks, for the performance evaluation.

\textbf{Evaluation protocol.} We report the Number of Clicks (NoC) measure which counts the average number of clicks required to achieve a target intersection over union (IoU) with ground truth mask. We set the target IoU score to 85\% or 90\% for different datasets, denoting the corresponding measures as NoC@85 and NoC@90 respectively. For a fair comparison, we use the same click generation strategy as in \cite{li2018interactive, xu2016deep} that operates as follows. It finds the dominant type of prediction errors (false positives or false negatives) and generates the next negative or positive click respectively at the point farthest from the boundaries of the corresponding error region.

\textbf{Network architecture.} In this work, we do not focus on network architecture improvements, so in all our experiments we use the standard DeepLabV3+ \cite{chen2018encoder} which is a state-of-the-art model for semantic segmentation. The architecture of our network is shown in Figure \ref{fig:f_brs}.

The model contains Distance Maps Fusion (DMF) block for adaptive fusion of RGB image and distance maps. It takes a concatenation of RGB image and 2 distance maps (one for positive clicks and one for negative clicks) as an input. The DMF block processes the 5-channel input with $1 \times 1$ convolutions followed by LeakyReLU and outputs a 3-channel tensor which can be passed into the backbone that was pre-trained on RGB images.

\begin{table}[t]
\small
\begin{center}
\begin{tabular}{l|c|c|c|c}
\hline
\multirow{2}{*}{Method} & \multicolumn{2}{c|}{GrabCut} & \multicolumn{2}{c}{Berkeley} \\
\cline{2-5}
 & w/o ZI & ZI & w/o ZI & ZI \\
\hline
\hline
Ours w/o BRS & 3.42 & 3.32 & 7.13 & 5.18 \\
Ours f-BRS-B & 2.68 & 2.98 & 5.69 & 4.34 \\
\hline
\end{tabular}
\end{center}
\caption{Evaluation of the proposed methods with ResNet-50 backbone with and without Zoom-In (ZI) on GrabCut and Berkeley datasets using NoC@90 (see Section \ref{sec:experiments}).}
\label{tab:comparison_zoomin}

\footnotesize
\begin{center}
\begin{tabular}{c|c|c|c|c}
\hline 
Data & Model & \begin{tabular}{@{}c@{}}\#images \\ $\geq$20 \end{tabular} & \begin{tabular}{@{}c@{}}\#images \\ $\geq$100 \end{tabular} & NoC\textsubscript{100}@90 \\
\hline 
\hline
\multirow{4}{*}{Berkeley} &   \cite{jang2019interactive} w/o BRS & 32 & 31 & 33.24 \\
 & \cite{jang2019interactive} BRS  & 10 & 2 & 8.77 \\
 & Ours w/o BRS  & 12 & 9 & 12.98 \\
 & Ours w f-BRS-B  & \textbf{2} & \textbf{0} & \textbf{4.47} \\
\hline
\multirow{4}{*}{DAVIS} & \cite{jang2019interactive} w/o BRS & 166 & 157 & 47.95 \\
 & \cite{jang2019interactive} BRS & \textbf{77} & 51 & 20.98 \\
 & Ours w/o BRS  & 92 & 81 & 27.58 \\
 & Ours w f-BRS-B  & 78 & \textbf{50} & \textbf{20.7} \\
\hline
\multirow{2}{*}{SBD}  & Ours w/o BRS  & 1650 & 1114 & 23.18 \\
& Ours w f-BRS-B  & 1466 & \textbf{265} & \textbf{14.98} \\
\hline
\end{tabular}
\end{center}
\caption{Convergence analysis on Berkeley, SBD and DAVIS datasets. We report the number of images that were not correctly segmented after 20 and 100 clicks and the NoC\textsubscript{100}@90 performance measure.}
\label{tab:evaluation_noc20_noc100}
\end{table}

\textbf{Implementation details.}
We formulate the training task as a binary segmentation problem and use normalized focal loss (NFL) introduced in \cite{sofiiuk2019adaptis} for training. We compare results of training with NFL and binary cross entropy in Appendix \ref{appendix:loss_ablation}.  We train all the models on image crops of size \mbox{$320 \times 480$} with horizontal and vertical flips as augmentations. We randomly resize images from $0.75$ to $1.25$ of original size before cropping. 

We sample clicks during training following the standard procedure first proposed in \cite{xu2016deep}. We set the maximum number of positive and negative clicks to 10, resulting in a maximum of 20 clicks per image.

In all experiments, we used Adam with $\beta_1=0.9$, $\beta_2=0.999$ and trained the networks for 120 epochs (100 epochs with learning rate $5 \times 10^{-4}$, last 20 epochs with learning rate $5 \times 10^{-5}$). The batch size was set to 28 and we used synchronous BatchNorm for all experiments. We trained ResNet-34 and ResNet-50 on 2 GPUs (Tesla P40) and ResNet-101 was trained on 4 GPUs (Tesla P40). The learning rate for the pre-trained ResNet backbone was 10 times lower than the learning rate for the rest of the network. We set the value of $\lambda$ to $10^{-3}$ for RGB-BRS and to $10^{-4}$ for all variations of f-BRS.

We use MXNet Gluon \cite{chen2015mxnet} with GluonCV \cite{he2018bag} framework for training and inference of our models. We take pre-trained models for ResNet-34, ResNet-50 and ResNet-101 from GluonCV Model Zoo.

\subsection{Convergence analysis}
\label{sec:convergence}

\begingroup
\begin{figure}
    \centering
    \includegraphics[width=\linewidth]{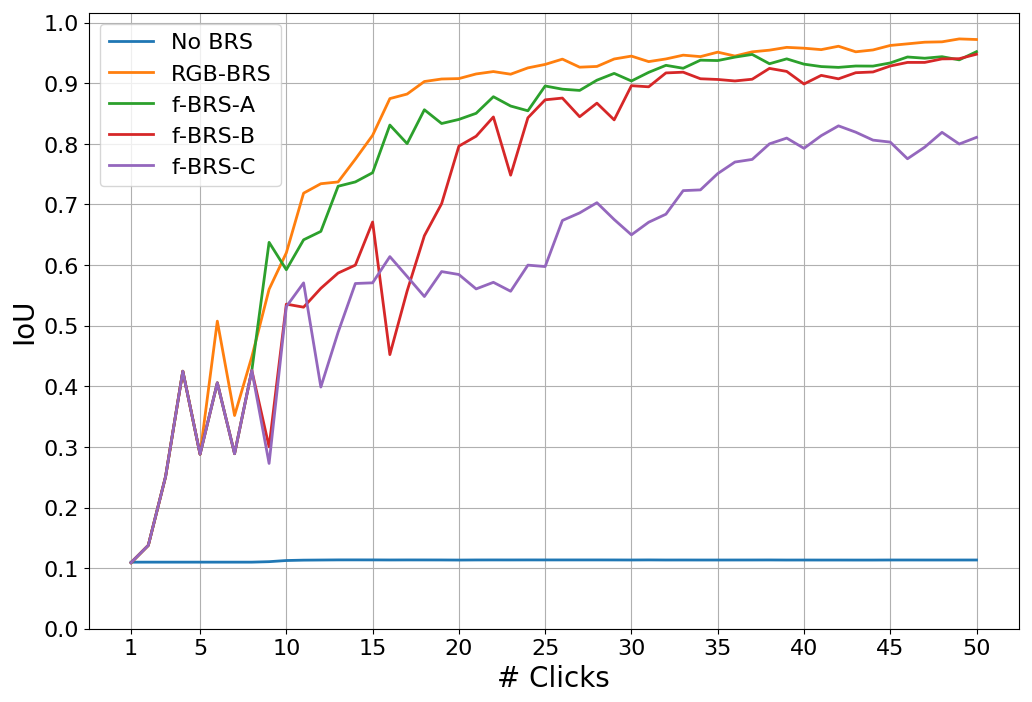}
    \caption{IoU with respect to the number of clicks added by a user for one of the most difficult image from GrabCut dataset (scissors). All results are obtained using the same model with ResNet-50. One can see that without BRS the model does not converge to the correct results. }
    \label{fig:scissors_no_convergence}
\end{figure}
\endgroup

An ideal interactive segmentation method should demonstrate predictable performance even for unseen object categories or unusual demand from the user. Moreover, the hard cases that require a significant amount of user input are the most interesting in the data annotation scenario. Thus, the desired property of an interactive segmentation method is convergence, \ie we expect the result to improve with adding more clicks and finally achieve satisfactory accuracy.

However, neither the training procedure nor the inference in feed-forward networks for interactive segmentation guarantee convergence. Therefore, we noticed that when using feed-forward networks, the result does not converge for a significant number of images, \ie additional user clicks do not improve the resulting segmentation mask. An example of such behaviour can be found in Figure  \ref{fig:scissors_no_convergence}. We observe very similar behaviour with different network architectures, namely with an architecture from \cite{jang2019interactive} and with DeepLabV3+. Below we describe our experiments. 

\textbf{Motivation for using NoC\textsubscript{100} metric.} Previous works usually report NoC with the maximum number of generated clicks limited to 20 (we simply call this metric NoC). However, for a large portion of images in the standard datasets, this limit is exceeded. In terms of NoC, images that require 20 clicks and 2000 clicks to obtain accurate masks will get the same penalty. Therefore, NoC does not distinguish between the cases where an interactive segmentation method requires slightly more user input to converge and the cases where it fails to converge (i.e. unable to achieve satisfactory results after any reasonable number of user clicks).

In the experiments below we analyse NoC with the maximum number of clicks limited to 100 (let us call this metric NoC\textsubscript{100}). NoC\textsubscript{100} is better for the convergence analysis, allowing us to identify the images where interactive segmentation fails. We believe that NoC\textsubscript{100} is substantially more adequate for comparison of interactive segmentation methods than NoC.

\begingroup
\begin{figure*}[!ht]
    \centering
    \includegraphics[width=0.9\linewidth]{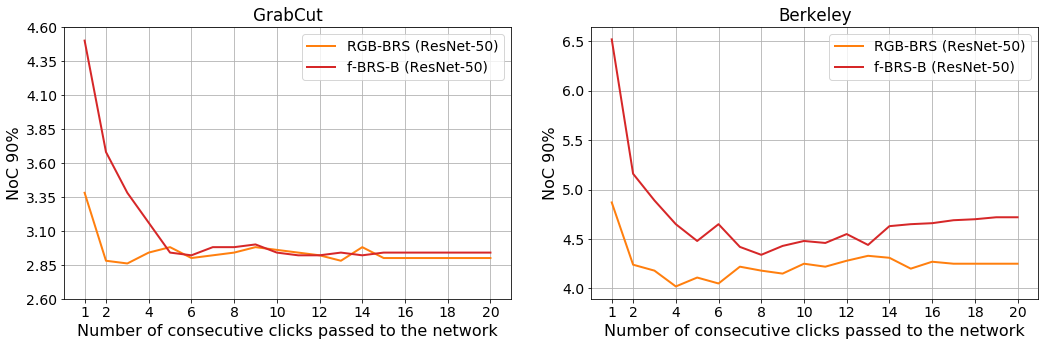}
    \caption{Evaluation of different click-processing strategies on GrabCut and Berkeley datasets. The plots show NoC@90 with respect to the number of consecutive clicks passed to the network.}
    \label{fig:clicks_limit_ablation}
\end{figure*}
\endgroup

\textbf{Experiments and discussion.} In Table \ref{tab:evaluation_noc20_noc100} we report the number of images that were not correctly segmented even after 20 and 100 clicks, and NoC\textsubscript{100} for the target IoU=90\% (NoC\textsubscript{100}@90).

One can see that both DeepLabV3+ and the network architecture from \cite{jang2019interactive} without BRS were unable to produce accurate segmentation results on a relatively large portion of images from all datasets even with 100 user clicks provided. Interestingly, this percentage is also high for the SBD dataset which has the closest distribution to the training set. The images that could not be segmented with 100 user clicks are clear failure cases for the method. The use of both original BRS and proposed f-BRS allows to reduce the number of such cases by several times and results in significant improvement in terms of NoC\textsubscript{100}. 

We believe that the use of optimization-based backpropagating refinement results not just in metrics improvement, but more importantly, it changes the behaviour of the interactive segmentation system and its convergence properties.

\subsection{Evaluation of the importance of clicks passed to the network}
We have noticed that the results do not always improve with the increasing number of clicks passed to the network.  Moreover, too many clicks can cause unpredictable behaviour of the network. On the other hand, the formulation of the optimization task for backpropagating refinement enforces the consistency of the resulting mask with user-provided annotation. 

One may notice that we can handle user clicks only as a target for BRS loss function without passing them to the network through distance maps. We initialise the state of the network by making a prediction with the first few clicks. Then we iteratively refine the resulting segmentation mask only with BRS according to the new clicks. 

We studied the relation between the number of consecutive clicks passed to the network and resulting NoC@90 on GrabCut and Berkeley datasets. The results of this study for RGB-BRS and f-BRS-B are shown in Figure \ref{fig:clicks_limit_ablation}. The results show that providing all clicks to the network is not an optimal strategy. It is clear that for RGB-BRS, the optimum is achieved by limiting the number of clicks to 4, and for f-BRS-B -- by 8 clicks. This shows that both BRS and f-BRS can adapt the network output to user input, without explicitly passing clicks to the network.

In all other experiments, we have limited the number of clicks passed to the network to 8 for f-BRS algorithms and to 4 for RGB-BRS.

\subsection{Comparison with previous works}

\begingroup
\begin{table*}[t]
\small
\begin{center}
\begin{tabular}{ll|c|c|c|c|c|c|c}
\hline
\multicolumn{2}{l|}{\multirow{2}{*}{Method}} & \multicolumn{2}{c|}{GrabCut} & Berkeley & \multicolumn{2}{c|}{SBD} & \multicolumn{2}{c}{DAVIS} \\
\cline{3-9}
\multicolumn{2}{l|}{} & NoC@85 & NoC@90 & NoC@90 & NoC@85 & NoC@90 & NoC@85 & NoC@90 \\
\hline
\hline
\multicolumn{2}{l|}{Graph cut \cite{boykov2001interactive}} & 7.98 & 10.00 & 14.22 & 13.6 & 15.96 & 15.13 & 17.41 \\
\multicolumn{2}{l|}{Geodesic matting \cite{gulshan2010geodesic}} & 13.32 & 14.57 & 15.96 & 15.36 & 17.60 & 18.59 & 19.50 \\
\multicolumn{2}{l|}{Random walker \cite{grady2006random}} & 11.36 & 13.77 & 14.02 & 12.22 & 15.04 & 16.71 & 18.31 \\
\multicolumn{2}{l|}{Euclidean star convexity \cite{gulshan2010geodesic}} & 7.24 & 9.20 & 12.11 & 12.21 & 14.86 & 15.41 & 17.70 \\
\multicolumn{2}{l|}{Geodesic star convexity \cite{gulshan2010geodesic}} & 7.10 & 9.12 & 12.57 & 12.69 & 15.31 & 15.35 & 17.52 \\
\multicolumn{2}{l|}{Growcut \cite{vezhnevets2005growcut}} & -- & 16.74 & 18.25 & -- & -- & -- & -- \\
\hline
\multicolumn{2}{l|}{DOS w/o GC \cite{xu2016deep}} & 8.02 & 12.59 & -- & 14.30 & 16.79 & 12.52 & 17.11 \\
\multicolumn{2}{l|}{DOS with GC \cite{xu2016deep}} & 5.08 & 6.08 & -- & 9.22 & 12.80 & 9.03 & 12.58 \\
\multicolumn{2}{l|}{Latent diversity \cite{li2018interactive}} & 3.20 & 4.79 & -- & 7.41 & 10.78 & \textbf{5.05} & 9.57 \\
\multicolumn{2}{l|}{RIS-Net \cite{liew2017regional}} & -- & 5.00 & -- & 6.03 & -- & -- & -- \\
\multicolumn{2}{l|}{CM guidance \cite{majumder2019content}} & -- & 3.58 & 5.60 & -- & -- & -- & --\\
\multicolumn{2}{l|}{BRS \cite{jang2019interactive}} & 2.60 & 3.60 & 5.08 & 6.59 & 9.78 & 5.58 & 8.24 \\
\hline
\hline
\multirow{3}{*}{Ours w/o BRS} & ResNet-34 & 2.52 & 3.20 & 5.31 & 5.51 & 8.58 & 5.47 & 8.51 \\
& ResNet-50 & 2.64 & 3.32 & 5.18 & 5.10 & \underline{8.01} & 5.39 & 8.18 \\
& ResNet-101 & 2.50 & 3.18 & 6.25 & 5.28 & 8.13 & \underline{5.12} & 8.01 \\
\hline
\multirow{3}{*}{Ours f-BRS-B} & ResNet-34 & \textbf{2.00} & \textbf{2.46} & 4.65 & 5.25 & 8.30 & 5.39 & 8.21 \\
& ResNet-50 & 2.50 & 2.98 & \textbf{4.34} & \underline{5.06} & 8.08 & 5.39 & \underline{7.81} \\
& ResNet-101 & \underline{2.30} & \underline{2.72} & \underline{4.57} & \textbf{4.81} & \textbf{7.73} & \textbf{5.04} & \textbf{7.41} \\
\hline
\end{tabular}
\end{center}
\caption{Evaluation results of GrabCut, Berkeley, SBD and DAVIS datasets. The best and the second best results are written in bold and underlined respectively.}
\label{tab:evaluation_cityscapes}

\small
\begin{center}
\begin{tabular}{l|c|c|c|c|c|c|c|c}
\hline
\multirow{2}{*}{Method} & \multicolumn{4}{c|}{Berkeley} & \multicolumn{4}{c}{Davis} \\
\cline{2-9}
& NoC@90 & \begin{tabular}{@{}c@{}}\#images \\ $\geq$20 \end{tabular} & SPC & Time, s & NoC@90 & \begin{tabular}{@{}c@{}}\#images \\ $\geq$20 \end{tabular} & SPC & Time, s \\
\hline
\hline
Ours w/o BRS & 5.18 & 12 & 0.091 & 49.9 & 8.18 & 92 & 0.21 & 585.9 \\
\hline
Ours RGB-BRS & 4.08 & 4 & 1.117 & 455.7 & 7.58 & 72 & 2.89 & 7480.8 \\
Ours DistMap-BRS & 4.17 & 4 & 0.669 & 276.4 & 7.93 & 73 & 1.47 & 4028.4 \\
Ours f-BRS-A & 4.36 & 3 & 0.281 & 119.3 & 7.54 & 72 & 0.75 & 1980.5 \\
Ours f-BRS-B & 4.34 & 2 & 0.132 & 55.07 & 7.81 & 78 & 0.32 & 889.4 \\
Ours f-BRS-C & 4.91 & 8 & 0.138 & 61.4 & 7.91 & 84 & 0.31 & 848.2 \\
\hline
\end{tabular}
\end{center}
\caption{Comparison of the results without BRS and with f-BRS types A, B and C with ResNet-50 backbone.}
\label{tab:comparison_fbrs}
\end{table*}
\endgroup

\textbf{Comparison using the standard protocol.} Table \ref{tab:evaluation_cityscapes} compares with previous works across the standard protocol and report the average NoC with two IoU thresholds: 85\% and 90\%. 

The proposed f-BRS algorithm requires fewer clicks than conventional algorithms, which indicates that the proposed algorithm yields accurate object masks with less user effort. 

We tested three backbones on all datasets. Surprisingly, there is no significant difference in performance between these models. The smallest ResNet-34 model shows the best quality on GrabCut dataset outperforming much heavier models such as ResNet-101. However, during training there was a significant difference in the values of the target loss function on the validation set between these models. This shows that the target loss function is poorly correlated with the NoC metric.

\textbf{Running time analysis.} We measure the average running time of the proposed algorithm in seconds per click (SPC) and measure the total running time to process a dataset. The SPC shows the delay between placing a click and getting the updated result. The second metric indicates the total time a user needs to spend to obtain a satisfactory image annotation. In these experiments, we set the threshold on the number of clicks per image to 20. We test it on Berkeley and DAVIS datasets using a PC with an AMD Ryzen Threadripper 1900X CPU and a GTX 1080 Ti GPU. 

Table \ref{tab:comparison_fbrs} shows the results for different versions of the proposed method and for our implemented baselines: without BRS and with RGB-BRS. The running time of f-BRS is an order of magnitude lower compared to RGB-BRS and adds just a small overhead with respect to a pure feed-forward model.

\subsection{Comparison of different versions of f-BRS.} 
The choice of a layer where to introduce auxiliary variables provides a trade-off between speed and accuracy of f-BRS. We compare three options of applying scale and bias to intermediate outputs in different parts of the network: after the backbone (f-BRS-A), before the first separable convolutions block in DeepLabV3+ (f-BRS-B), and before the second separable convolutions block in DeepLabV3+ (f-BRS-C). As a baseline for our experiments, we report the results for a feed-forward network without BRS. We also implement RGB-BRS, employing the optimization with respect to an input image. In these experiments, we used the ResNet-50 backbone.

We report NoC@90 and the number of images where the satisfactory result was not obtained after 20 user clicks. We also measure SPC (seconds per click) and Time (total time to process a dataset). Notice that direct comparison of the timings with the numbers reported in previous works is not valid due to differences in used frameworks and hardware. Therefore, only relative comparison makes sense. 

The results of the evaluation for Berkeley and DAVIS datasets are shown in Table \ref{tab:comparison_fbrs}. One can notice that all versions of f-BRS perform better than the baseline without BRS. The f-BRS-B is about 8 times faster than the RGB-BRS while showing very close results in terms of NoC. Therefore, we chose it for the comparative experiments.

\section{Conclusions}

We proposed a novel backpropagating refinement scheme (f-BRS) that operates on intermediate features in the network and requires running forward and backward passes just for a small part of a network. Our approach was evaluated on four standard interactive segmentation benchmarks and set new state-of-the-art results in terms of both accuracy and speed. The conducted experiments demonstrated a better convergence of backpropagating refinement schemes compared to pure feed-forward approaches. We analysed the importance of first clicks passed to the network and showed that both BRS and f-BRS can successfully adapt the network output to user input, without explicitly passing clicks to the network.

{\small
\bibliographystyle{ieee_fullname}
\bibliography{egbib}
}

\addtocontents{toc}{\protect\contentsline{section}{Appendix:}{}}
\newpage
\clearpage
\begin{appendices}

\section{Analysis of the average IoU according to the number of clicks}

We computed the mean IoU score according to the number of clicks for GrabCut, Berkeley, SBD and DAVIS datasets (see Figure \ref{fig:iou_ablation}). We also evaluated the BRS~\cite{jang2019interactive}~model from authors' public repository for a fair comparison. 

On the plots you can see that f-BRS-B has drops on DAVIS and SBD datasets at the number of clicks 9. This is due to the fact that f-BRS can sometimes fall into a bad local minimum. This issue can be solved by setting a higher regularization coefficient $\lambda$ in the BRS loss function. However, with the increase of the $\lambda$, the convergence of the method at a large number of clicks becomes worse.

\section{Measuring the limitation of f-BRS}
\label{sec:oracle}

\begin{figure*}
\begin{center}
\begin{tabular}{cccc}
    \includegraphics[width=0.42\linewidth]{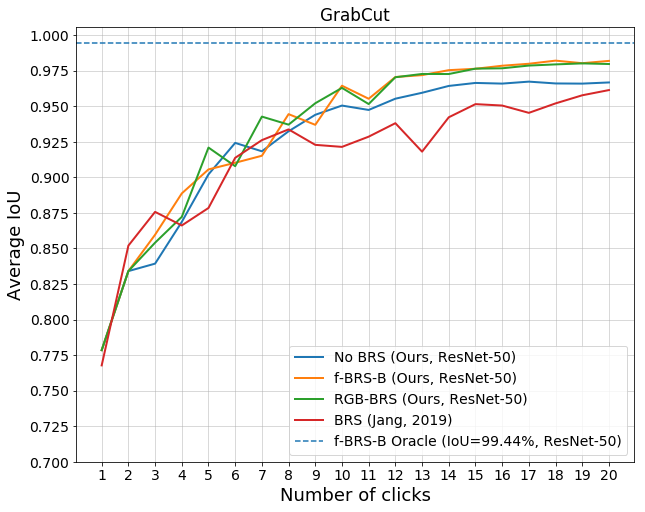} &
    \includegraphics[width=0.42\linewidth]{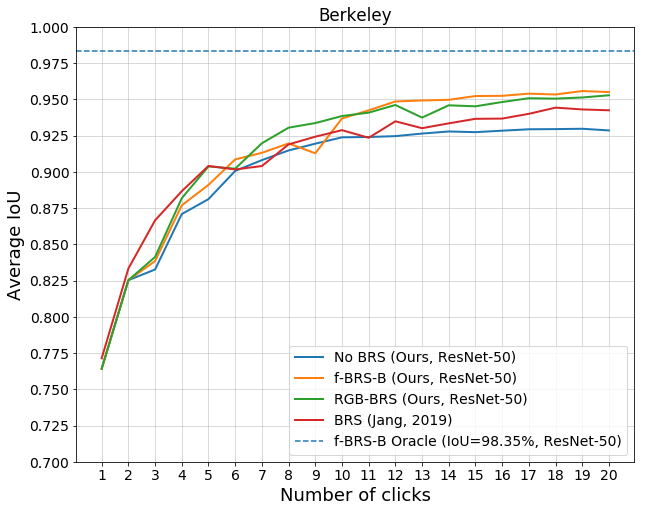} \\
    \includegraphics[width=0.42\linewidth]{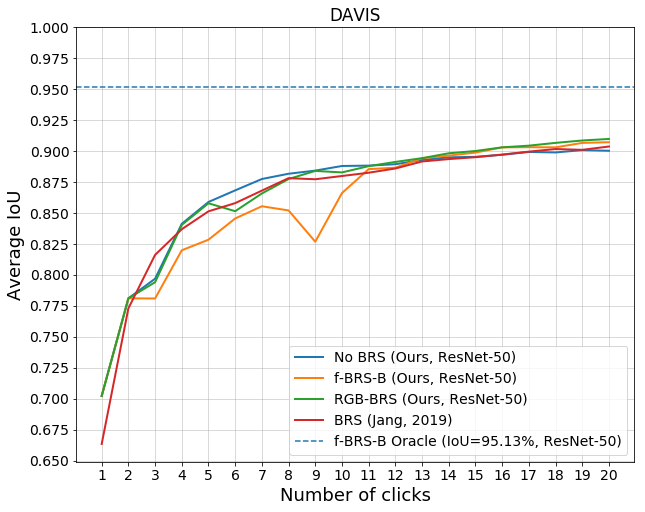} &
    \includegraphics[width=0.42\linewidth]{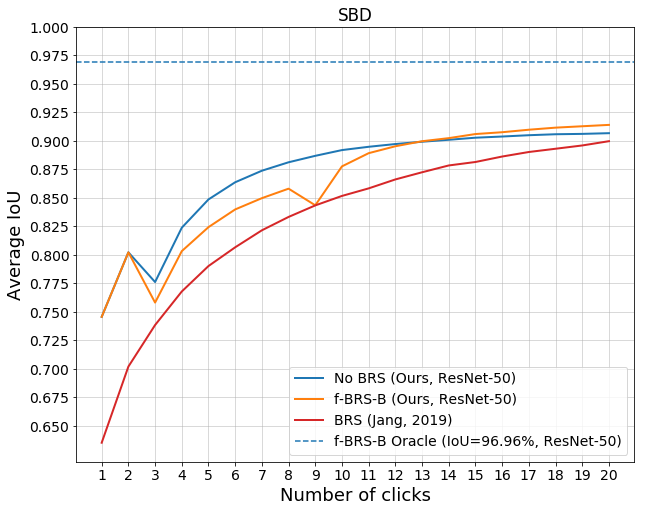} \\
\end{tabular}
\end{center}
\caption{Comparison of the average IoU scores according to the number of clicks on GrabCut, Berkeley, DAVIS and SBD datasets. The dashed horizontal line shows the average IoU limit that can theoretically be reached by f-BRS-B method (for more details see Section~\ref{sec:oracle}).}
\label{fig:iou_ablation}
\end{figure*}

We decided to find out the limit of accuracy that can be obtained using only f-BRS, adjusting scales and biases for an intermediate layer in the DeepLabV3+ head. For this, we first evaluated the model for 20 clicks using the standard protocol. Then we continued with L-BFGS-B optimization for scales and biases using ground truth mask as loss target instead of interactive clicks. It equals to using all pixels of the image as input clicks (positive click for each foreground pixel and negative for each background pixel). We estimated the mean IoU score for each dataset which is shown in Figure \ref{fig:iou_ablation} (f-BRS-B Oracle).

The figure illustrates that the accuracy limit the algorithm can reach is highly dependent on the dataset. DAVIS and SBD datasets are much harder than GrabCut and Berkeley. DAVIS has many complex masks labeled with pixel perfect precision, which is closer to the task of image matting. On the contrary, SBD has many masks with rough or inaccurate annotation.

\section{Full evaluation results for all our methods}

We report the NoC@85 and NoC@90 metrics for GrabCut, Berkeley, SBD and DAVIS datasets for all BRS variations with different backbones (ResNet-34, ResNet-50 and ResNet-101). The use of BRS leads to consistent improvement in accuracy. All these results are presented in Table~\ref{tab:evaluation_all}.

Overall, the choice of a backbone only slightly affects the methods' accuracy on GrabCut and Berkeley datasets. However, we noticed a significant difference between ResNet-34 and ResNet-101 while testing on SBD validation dataset, which has the closest distribution to the training one. In most cases, DistMap-BRS shows slightly worse NoC compared to RGB-BRS. 

\setlength{\tabcolsep}{3.0pt} % Default value: 6pt
\renewcommand{\arraystretch}{0.7} % Default value: 1
\setlength{\fboxsep}{0pt}%
\setlength{\fboxrule}{1pt}%

{
\renewcommand{\arraystretch}{1.0}
\begin{table*}[ht]
\small
\begin{center}
\begin{tabular}{ll|c|c|c|c|c|c|c|c}
\hline
\multicolumn{2}{l|}{\multirow{2}{*}{Method}} & \multicolumn{2}{c|}{GrabCut} & \multicolumn{2}{c|}{Berkeley} & \multicolumn{2}{c|}{SBD} & \multicolumn{2}{c}{DAVIS} \\
\cline{3-10}
\multicolumn{2}{l|}{} & NoC@85 & NoC@90 & NoC@85 & NoC@90 & NoC@85 & NoC@90 & NoC@85 & NoC@90 \\
\hline
\hline
\multirow{3}{*}{Ours w/o BRS} & ResNet-34 & 2.52 & 3.20 & 3.09 & 5.31 & 5.51 & 8.58 & 5.47 & 8.51 \\
& ResNet-50 & 2.64 & 3.32 & 3.29 & 5.18 & 5.10 & 8.01 & 5.39 & 8.18 \\
& ResNet-101 & 2.50 & 3.18 & 3.45 & 6.25 & 5.28 & 8.13 & 5.12 & 8.01 \\
\hline
\multirow{3}{*}{Ours RGB-BRS} & ResNet-34 & 2.00 & 2.52 & 2.51 & 4.28 & 4.72 & 7.45 & 5.30 & 7.86 \\
& ResNet-50 & 2.38 & 2.94 & 2.65 & 4.08 & 4.45 & 7.12 & 5.28 & 7.58 \\
& ResNet-101 & 2.00 & 2.48 & 2.26 & 4.21 & 4.17 & 6.69 & 4.95 & 7.09 \\
\hline
\multirow{3}{*}{Ours DistMap-BRS} & ResNet-34 & 1.98 & 2.54 & 2.45 & 4.41 & 4.85 & 7.66 & 5.34 & 8.11 \\
& ResNet-50 & 2.36 & 2.90 & 2.67 & 4.17 & 4.63 & 7.37 & 5.35 & 7.93 \\
& ResNet-101 & 2.00 & 2.46 & 2.21 & 4.41 & 4.42 & 7.10 & 5.03 & 7.63 \\
\hline
\multirow{3}{*}{Ours f-BRS-A} & ResNet-34 & 1.94 & 2.54 & 2.66 & 4.36 & 5.11 & 8.17 & 5.39 & 8.09 \\
& ResNet-50 & 2.54 & 3.06 & 2.74 & 4.44 & 4.94 & 7.97 & 5.37 & 7.54 \\
& ResNet-101 & 2.08 & 2.62 & 2.39 & 4.79 & 4.68 & 7.58 & 5.01 & 7.21 \\
\hline
\multirow{3}{*}{Ours f-BRS-B} & ResNet-34 & 2.00 & 2.46 & 2.60 & 4.65 & 5.25 & 8.30 & 5.39 & 8.21 \\
& ResNet-50 & 2.50 & 2.98 & 2.77 & 4.34 & 5.06 & 8.08 & 5.39 & 7.81 \\
& ResNet-101 & 2.30 & 2.72 & 2.52 & 4.57 & 4.81 & 7.73 & 5.04 & 7.41 \\
\hline
\multirow{3}{*}{Ours f-BRS-C} & ResNet-34 & 2.10 & 2.54 & 2.72 & 4.48 & 5.23 & 8.11 & 5.47 & 8.35 \\
& ResNet-50 & 2.60 & 3.10 & 2.89 & 4.90 & 5.05 & 7.97 & 5.50 & 7.90 \\
& ResNet-101 & 2.18 & 2.68 & 2.64 & 4.64 & 4.85 & 7.64 & 5.11 & 7.37 \\
\hline
\end{tabular}
\end{center}
\caption{Evaluation results on GrabCut, Berkeley, SBD and DAVIS datasets.}
\label{tab:evaluation_all}
\end{table*}
}

\section{Additional interactive segmentation results}

\begin{figure*}
\includegraphics[width=\linewidth]{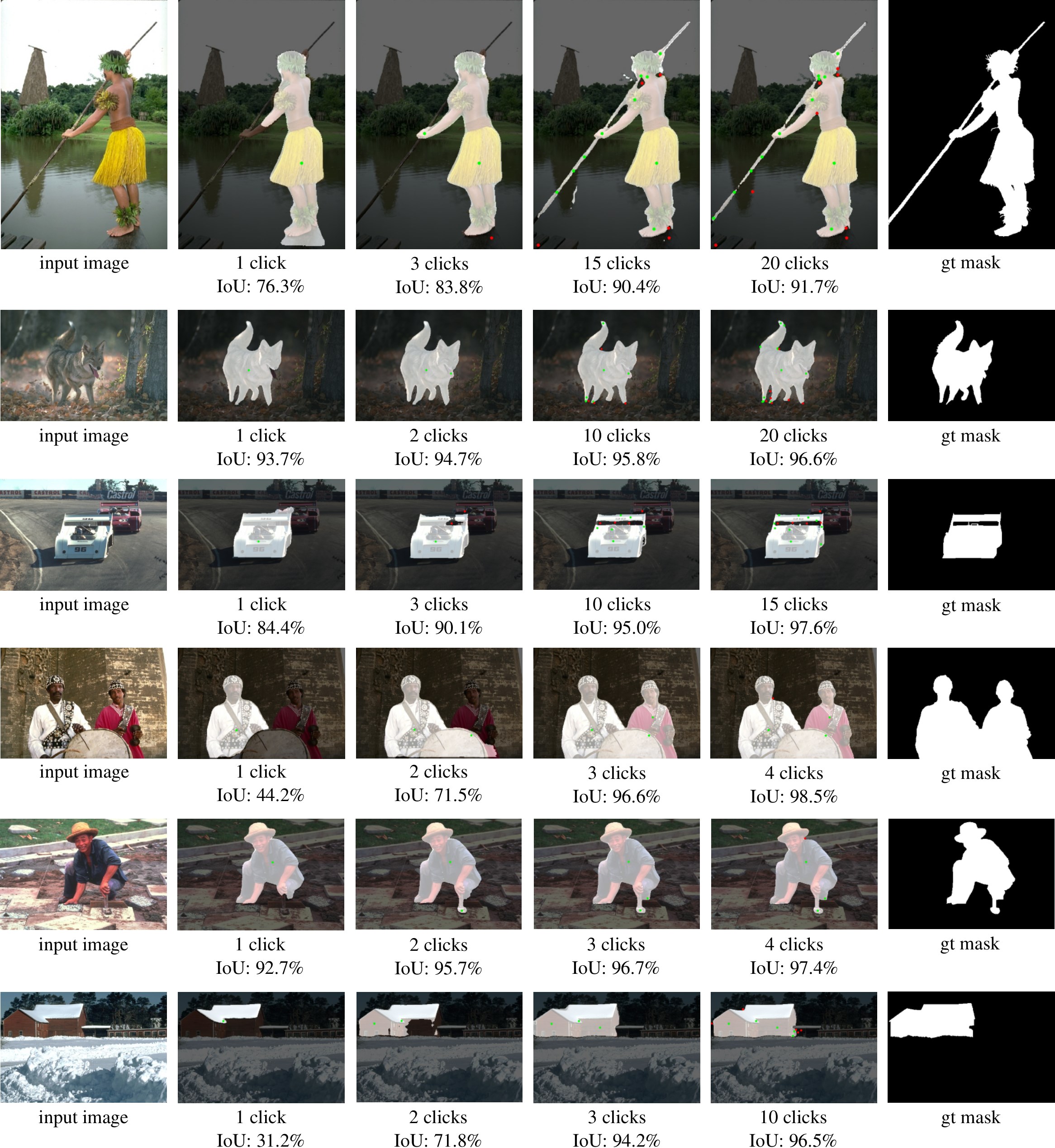}
\caption{Examples of good convergence of the proposed f-BRS-B method with ResNet-50 backbone on Berkeley dataset.}
\label{fig:good_results_1}
\end{figure*}

\begin{figure*}
\includegraphics[width=\linewidth]{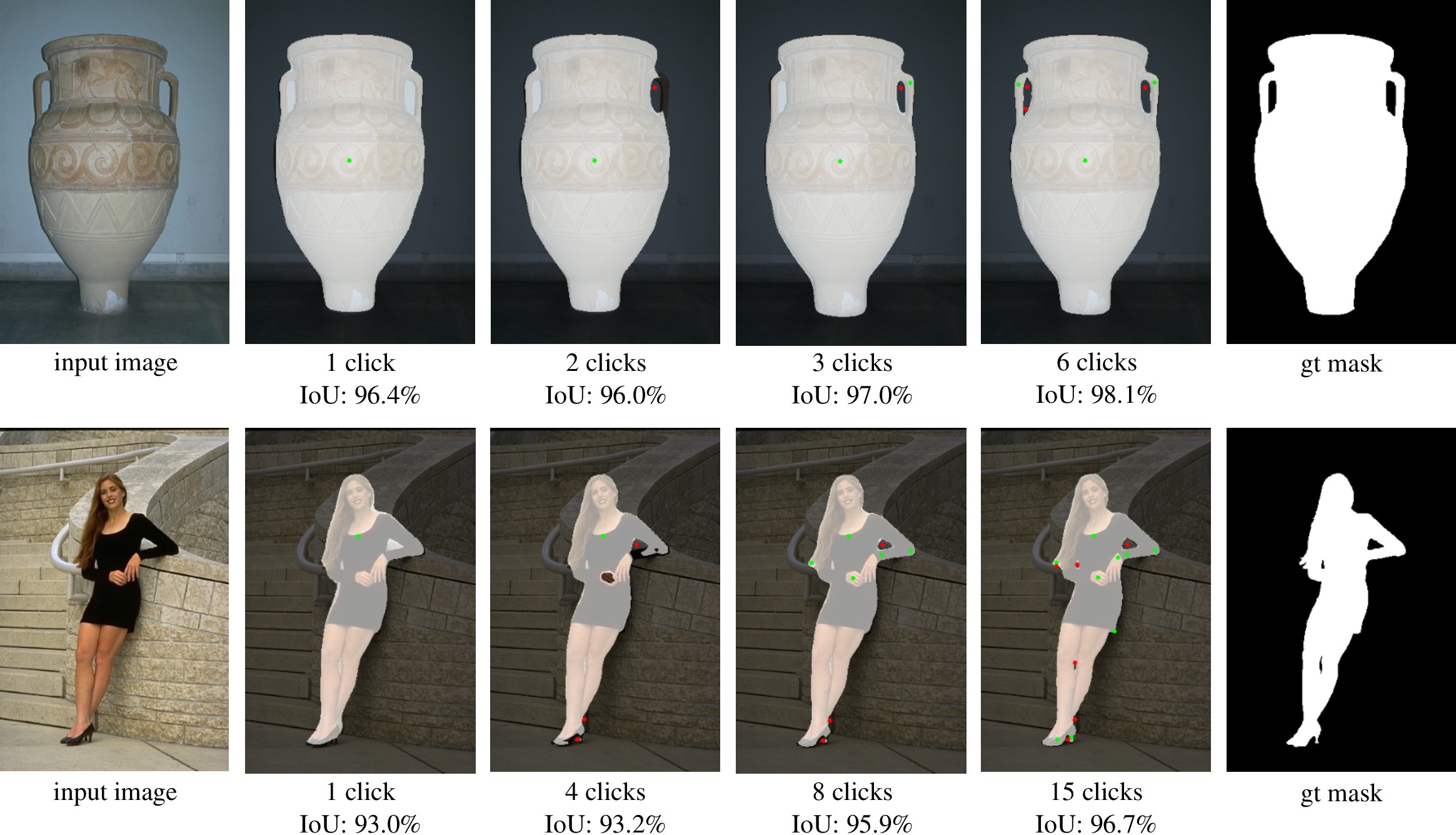}
\caption{Examples of good convergence of the proposed f-BRS-B method with ResNet-50 backbone on Berkeley dataset.}
\label{fig:good_results_2}
\end{figure*}

\begin{figure*}
\includegraphics[width=\linewidth]{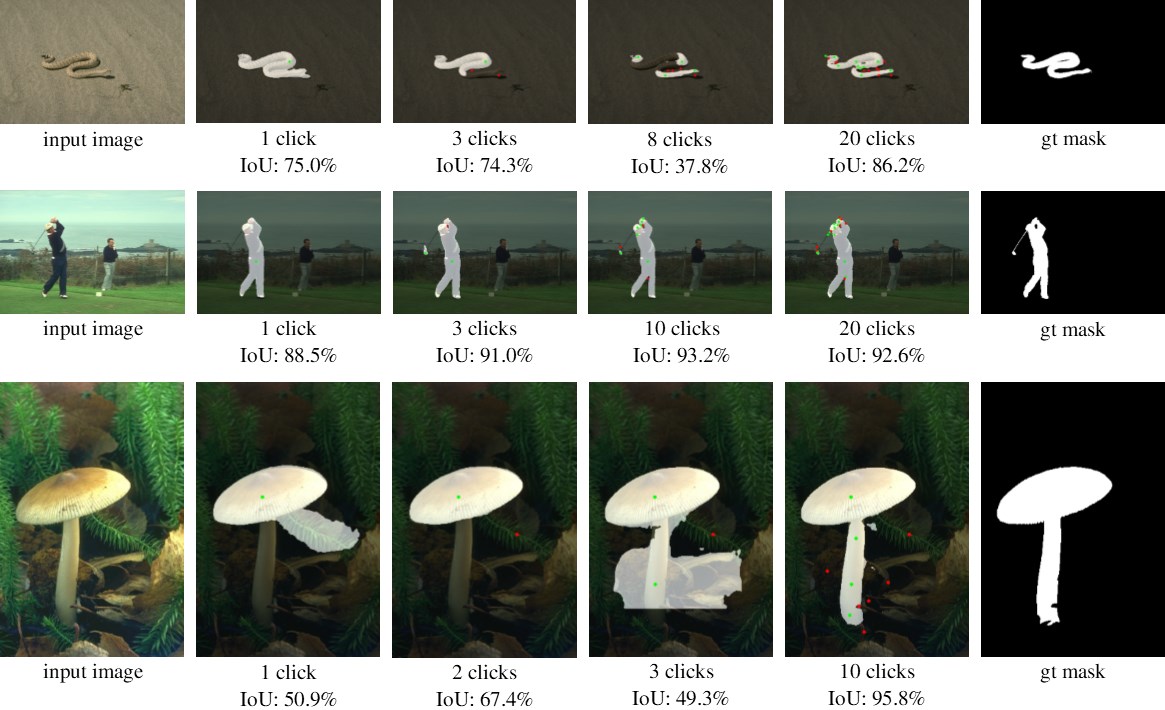}
\caption{Some challenging examples from Berkeley dataset.}
\label{fig:bad_results}
\end{figure*}

We also provide more results of our interactive segmentation algorithm (f-BRS-B with ResNet-50) on different images. Figure \ref{fig:good_results_1} and \ref{fig:good_results_2} represent good cases, while Figure \ref{fig:bad_results}  represents bad cases when testing on Berkeley dataset. 

\begin{figure*}
\centering
\includegraphics[width=\linewidth]{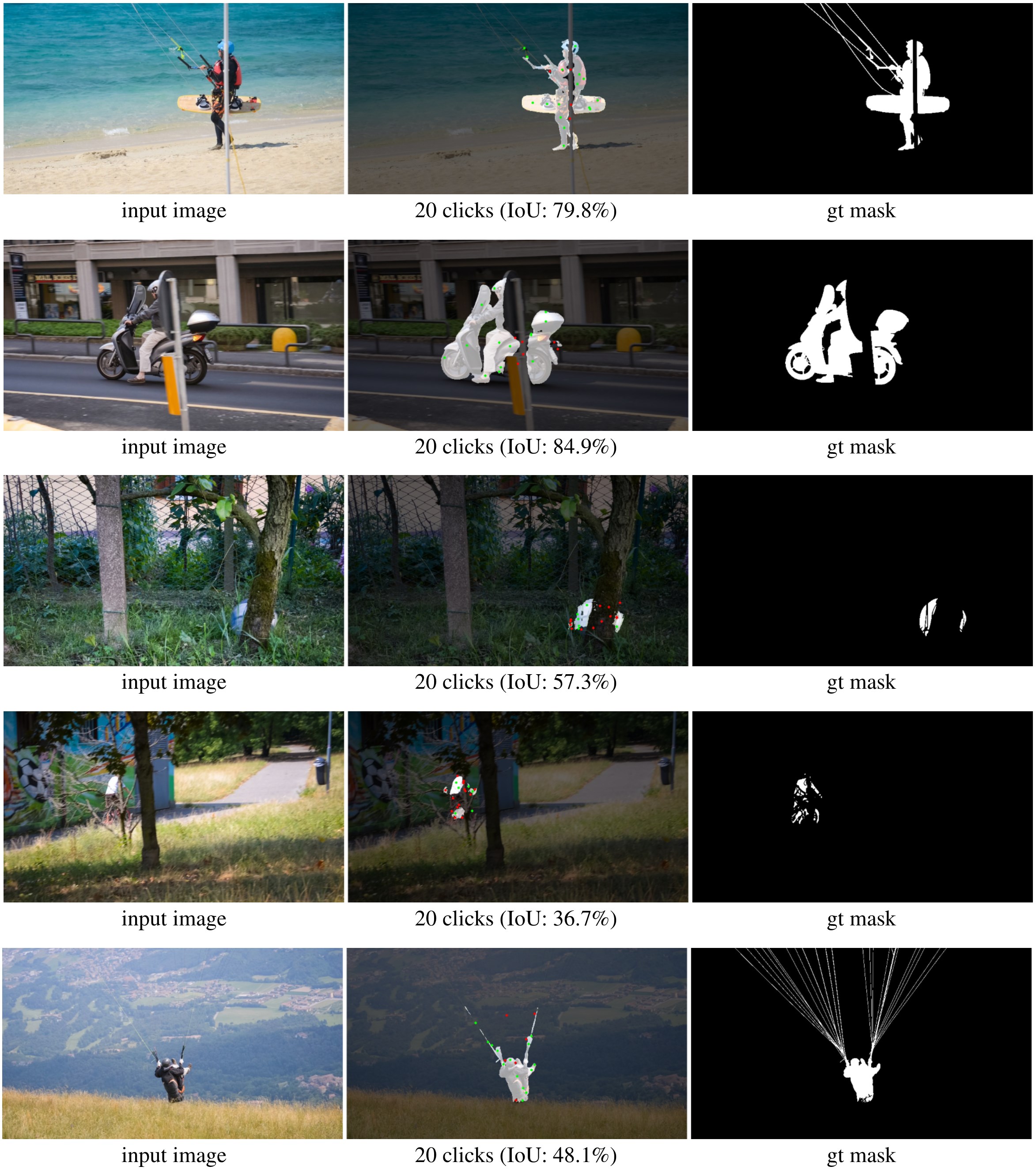}
\caption{Some of the worst examples from DAVIS dataset.}
\label{fig:davis_bad_results}
\end{figure*}

Figure \ref{fig:davis_bad_results} shows  some of the worst results of testing on DAVIS dataset. The algorithm does not even match 85\% IoU in 20 clicks.

\section{Loss function ablation}
\label{appendix:loss_ablation}

We use normalized focal loss (NFL) introduced in \cite{sofiiuk2019adaptis} as an alternative to binary cross entropy (BCE) in our experiments. NFL retains the advantages of focal loss \cite{lin2017focal} and allows to concentrate the training process on erroneous regions and at the same time, the total gradient of NFL does not fade over time and remains equal to the total gradient of BCE. Thus training with NFL leads to faster convergence and better accuracy compared to training with BCE. We provide an ablation study for all our models trained with NFL and BCE loss functions in Table \ref{tab:evaluation_cel_nfl} (NFL\textsuperscript{*} denotes results that were obtained with the latest code from our public GitHub repository \footnote{\url{https://github.com/saic-vul/fbrs_interactive_segmentation}} with minor technical improvements compared to the original code).

{
\renewcommand{\arraystretch}{1.2}
\begin{table*}[ht]
\begin{center}
\begin{tabular}{c|ll|c|c|c|c|c|c|c}
\hline
\multirow{2}{*}{Loss} & \multicolumn{2}{l|}{\multirow{2}{*}{Method}} & \multicolumn{2}{c|}{GrabCut} & Berkeley & \multicolumn{2}{c|}{SBD} & \multicolumn{2}{c}{DAVIS} \\
\cline{4-10}
& \multicolumn{2}{l|}{} & NoC@85 & NoC@90 & NoC@90 & NoC@85 & NoC@90 & NoC@85 & NoC@90 \\
\hline
\hline
\multirow{6}{*}{NFL} & \multirow{3}{*}{Ours w/o BRS} & ResNet-34 & 2.52 & 3.20 & 5.31 & 5.51 & 8.58 & 5.47 & 8.51 \\
& & ResNet-50 & 2.64 & 3.32 & 5.18 & 5.10 & 8.01 & 5.39 & 8.18 \\
& & ResNet-101 & 2.50 & 3.18 & 6.25 & 5.28 & 8.13 & 5.12 & 8.01 \\
\cline{2-10}
& \multirow{3}{*}{Ours f-BRS-B} & ResNet-34 & 2.06 & 2.48 & 4.17 & 4.47 & 7.28 & 5.34 & 7.73 \\
& & ResNet-50 & 2.20 & 2.64 & 4.22 & 4.55 & 7.45 & 5.44 & 7.81 \\
& & ResNet-101 & 2.30 & 2.68 & 4.22 & 4.20 & 7.10 & 5.32 & 7.35 \\
\hline
\hline
\multirow{6}{*}{NFL\textsuperscript{*}} & \multirow{3}{*}{Ours w/o BRS} & ResNet-34 & 2.44 & 3.18 & 4.91 & 4.77 & 7.57 & 5.46 & 8.14 \\
& & ResNet-50 & 2.24 & 2.74 & 6.17 & 5.06 & 7.99 & 5.51 & 8.31 \\
& & ResNet-101 & 2.84 & 3.18 & 5.36 & 4.96 & 7.96 & 5.32 & 8.07 \\
\cline{2-10}
& \multirow{3}{*}{Ours f-BRS-B} & ResNet-34 & 2.00 & 2.46 & 4.65 & 5.25 & 8.30 & 5.39 & 8.21 \\
& & ResNet-50 & 2.50 & 2.98 & 4.34 & 5.06 & 8.08 & 5.39 & 7.81 \\
& & ResNet-101 & 2.30 & 2.72 & 4.57 & 4.81 & 7.73 & 5.04 & 7.41 \\
\hline
\hline
\multirow{6}{*}{CEL} & \multirow{3}{*}{Ours w/o BRS} & ResNet-34 & 1.94 & 2.90 & 5.52 & 5.51 & 8.55 & 5.71 & 8.58 \\
& & ResNet-50 & 2.22 & 2.64 & 5.61 & 5.11 & 8.02 & 5.64 & 8.58 \\
& & ResNet-101 & 2.00 & 2.44 & 4.95 & 5.25 & 8.31 & 5.39 & 8.26 \\
\cline{2-10}
& \multirow{3}{*}{Ours f-BRS-B} & ResNet-34 & 2.00 & 2.72 & 4.87 & 5.02 & 8.02 & 5.89 & 8.70 \\
& & ResNet-50 & 2.24 & 2.72 & 4.67 & 4.81 & 7.61 & 5.76 & 8.64 \\
& & ResNet-101 & 2.10 & 2.40 & 4.12 & 4.81 & 7.68 & 5.57 & 8.18 \\
\end{tabular}
\end{center}
\caption{Comparison between NFL and CEL losses on GrabCut, Berkeley, SBD and DAVIS datasets.}
\label{tab:evaluation_cel_nfl}
\end{table*}
}

\end{appendices}
\end{document}